\documentclass[twoside]{article}

\usepackage[accepted]{aistats2020}
\usepackage{microtype}
\usepackage{graphicx}
\usepackage{tabularx,ragged2e,caption}
\usepackage{esvect}
\usepackage{booktabs} % for professional tables
\usepackage{amsmath, amssymb}
\usepackage{bm}
\usepackage{amsthm}

\newtheorem{theorem}{Theorem}

\usepackage{multirow}
\usepackage{diagbox}
\usepackage{enumitem}
\usepackage{mathtools}
\usepackage{graphicx}
\usepackage{subfig}
\usepackage{wrapfig}
\usepackage{esvect}
\usepackage{xcolor}
\usepackage{algorithm}
\usepackage{algorithmic}
\usepackage{wrapfig}
\usepackage{slashbox}
\usepackage{balance}

\newcommand{\Var}{\mathop{\mathrm{Var}}}

\newcommand{\KL}{\mathop{\mathrm{KL}}}

\usepackage{textcomp}

\usepackage{ragged2e}

\newcommand{\quo}[1]{\big[#1\big]}
\newcommand{\quooo}[1]{\Big[#1\Big]}

\newcommand{\indep}{\raisebox{0.08em}{\rotatebox[origin=c]{90}{$\models$}}}
\newcolumntype{C}[1]{>{\centering\arraybackslash}p{#1}}
\newcommand{\notindep}{\not\!\perp\!\!\!\perp}

\usepackage[utf8]{inputenc} % allow utf-8 input
\usepackage[T1]{fontenc}    % use 8-bit T1 fonts
\usepackage{hyperref}       % hyperlinks
\usepackage{url}            % simple URL typesetting
\usepackage{booktabs}       % professional-quality tables
\usepackage{amsfonts}       % blackboard math symbols
\usepackage{nicefrac}       % compact symbols for 1/2, etc.
\usepackage{microtype}      % microtypography
\usepackage{pifont}

\begin{document}

% If your paper is accepted and the title of your paper is very long,
% the style will print as headings an error message. Use the following
% command to supply a shorter title of your paper so that it can be
% used as headings.
%
%\runningtitle{I use this title instead because the last one was very long}

% If your paper is accepted and the number of authors is large, the
% style will print as headings an error message. Use the following
% command to supply a shorter version of the authors names so that
% they can be used as headings (for example, use only the surnames)
%
%\runningauthor{Surname 1, Surname 2, Surname 3, ...., Surname n}

\twocolumn[

\aistatstitle{Learning Overlapping Representations for the Estimation of Individualized Treatment Effects}

\aistatsauthor{ Yao Zhang \And Alexis Bellot \And  Mihaela van der Schaar }

\aistatsaddress{ University of Cambridge \And University of Cambridge \\ The Alan Turing Institute \And University of Cambridge, UCLA \\ The Alan Turing Institute } ]

\begin{abstract}
The choice of making an intervention depends on its potential benefit or harm in comparison to alternatives. Estimating the likely outcome of alternatives from observational data is a challenging problem as \textit{all} outcomes are never observed, and selection bias precludes the direct comparison of differently intervened groups. Despite their empirical success, we show that algorithms that learn domain-invariant representations of inputs (on which to make predictions) are often inappropriate, and develop generalization bounds that demonstrate the dependence on domain overlap and highlight the need for invertible latent maps. Based on these results, we develop a deep kernel regression algorithm and posterior regularization framework that substantially outperforms the state-of-the-art on a variety of benchmarks data sets.
\end{abstract}

\section{Introduction}
Counterfactual estimation poses the question of what would have been the outcome, if a different intervention had been applied. In order to make decisions in complex domains, making predictions on the causal effects of different actions and how these may vary across individuals is critical. In this paper we focus on the problem of making these predictions based on observational data, which is increasingly available in many domains such as medicine, public policy and advertising. In this setting, past actions, outcomes and context are available, but not knowledge of the treatment assignment policy - we do not know why a given individual was intervened on or not. The treatment assignment mechanism will often be causally affected by context variables that also causally influence the outcome. As an example, motivated unemployed individuals are more likely to both take advantage of government training programs \textit{and} find a new job soon. 

Learning from observational data requires adjusting for the covariate shift that exists between groups of individuals that are observed to have received different interventions. The challenge is how to untangle confounding factors and make valid counterfactual predictions - the response if a different treatment had been applied. Recent methods have predominantly focused on learning representations regularized to balance these confounding factors by enforcing domain invariance with distributional distances \cite{johansson2016learning,johansson2018learning,yao2018representation}. In this paper we argue that domain invariance is often too strict a requirement; overlapping support is sufficient for identifiability of the causal effect and equality in densities is not necessary. We interpret the loss in predictive power of domain-invariant representations as loss of information in the input variables that causally influence the treatment assignment, which is also often highly predictive of the treatment effect. Consider the example above for illustration: it is precisely because motivation is predictive of job outcomes that it confounds the treatment effect.

% \newcolumntype{A}{>{\centering\arraybackslash}m{0.5cm}}
% \newcolumntype{B}{>{\centering\arraybackslash}m{3.5cm}}
% \newcolumntype{C}{>{\centering\arraybackslash}m{3cm}}
% \newcolumntype{D}{>{\centering\arraybackslash}m{2.2cm}}
% \newcolumntype{E}{>{\centering\arraybackslash}m{2.9cm}}
% \newcolumntype{F}{>{\centering\arraybackslash}m{2.8cm}}

% \setlength\tabcolsep{4.6pt}
% \begin{table*}[t]
% \caption{Comparison of related theoretical works in the context of estimating individual treatment effect (ITE). Distribution distance between the treated and control group are mentioned in all the works. Our work additionally show that minimizing counterfactual variance and preserving input information content can lead to better generalization in ITE estimation.}
% \vspace{-1em}
% \begin{small}
% \begin{center}
% \begin{tabular}{ABCDEF}
% \toprule
%      Paper & {Model} & {Theory } & {\scriptsize Distribution Distance} & {\scriptsize Counterfactual Variance} & {\scriptsize Preserving Input Information Content } \\
% \midrule
% \cite{shalit2017estimating} & Neural Network & {\scriptsize Generalization Bound}  & \cmark & \xmark & \xmark \\
% \midrule
% \cite{alaa2018limits}    & Gaussian Process &  {\scriptsize Minimax Risk}  & \cmark & \xmark  & \xmark \\
% \midrule
% Ours & Deep Kernel Learning  &  {\scriptsize PAC-Bayes Bound} & \cmark & \cmark & \cmark \\
% \bottomrule
% \end{tabular}
% \vspace{-1.5em}
% \end{center}
% \label{tab:related}
% \end{small}
% \end{table*}

\begin{figure*}[t]
\centering
 \includegraphics[width=1\textwidth]{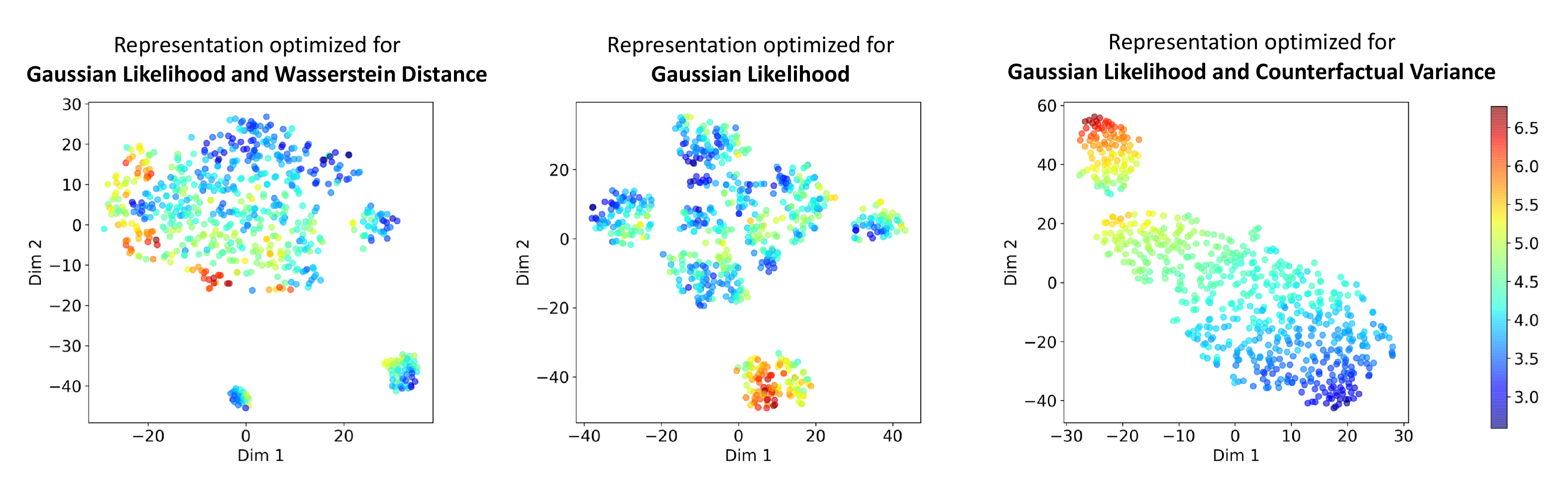}
\caption{T-SNE visualizations of the learned embeddings for the control potential outcomes $Y(0)$ of the IHDP dataset. Each panel shows representations regularized by different criteria and the colored heatmap represents different outcome magnitudes with different colors. The \textbf{left panel} shows representations regularized by the Wasserstein distributional distance and results in poor discrimination. The \textbf{middle panel} shows representations optimized only for the factual data with the Gaussian likelihood. The \textbf{right panel} shows representations regularized by the counterfactual variance, our proposed criteria. Much better separation in outcomes is obtained by regularizing for the predictive variance, in contrast to using integral probability metrics such as the Wasserstein distance. Similar plots for a comparison between the control and treated group, and the treated potential outcome $Y(1)$ can be found in Appendix 5.3.}
\label{fig:embedding}
\end{figure*}

We introduce an optimization framework based on regularizing posterior distributions of the treatment effect that includes existing representation learning algorithms for different choices of regularization terms. We take advantage of this framework to introduce a novel type of regularization criteria for the problem of treatment effect estimation: the posterior counterfactual variance for enforcing domain overlap, and invertible representations to preserve the information content of the underlying context. Such an objective enjoys much better generalization in small sample regimes, smoother representation surfaces with respect to the outcomes, as can be seen in Figure \ref{fig:embedding}, and a Bayesian treatment of paramaters which allows consistent uncertainty estimation in predictions. 

We summarize our \textbf{contributions} as follows:
\begin{enumerate}[topsep=0pt,itemsep=0pt]
    \item We develop a theory that justifies regularizing for the posterior variance to improve generalization error and establish the limitations of distributional distances.
    \item We propose to use deep kernels and posterior regularization as a general framework to learn individualized treatment effects for arbitrary regularization terms. 
    \item We provide an instantiation of this algorithm informed by novel generalization bounds that substantially improves upon the performance of state-of-the-art prediction algorithms. 
\end{enumerate}

\subsection{Related work} 
Due to the ability of deep neural nets to learn rich representations, recent advances in predicting individualized treatment effects have focused on learning representations invariant to the treatment assignment policy that achieve a small error on the factual data. The hope is that the learnt representation and prediction function can generalize to prediction of counterfactual outcomes. Several methods follow this approach. \cite{johansson2016learning} proposed learning a representation of the data that makes the treated and control distributions more similar, fitting a linear ridge-regression model on top of it. \cite{shalit2017estimating} built on their approach to derive a more flexible family of algorithms including non-linear hypotheses. However, both algorithms insist on quantifying divergence between treated and control groups with integral probability metrics. 

In this work, we share the need for good representations but argue for enforcing support overlap rather than equality in densities. \cite{yao2018representation}, inspired by nearest-neighbour methods, learn representation that preserve local similarity information in feature space and were able to show a decrease in the generalization error in counterfactual estimation.

Adapting Bayesian algorithms for the problem of individualized treatment effects has attracted a lot of interest, in particular in the field of medicine where quantifying uncertainty is important. \cite{alaa2017bayesian} regularize counterfactual predictions through their posterior variance and similarly stress the importance to provide confidence in their estimates using credible intervals, but did not investigate the generalization properties of their algorithm and only allowed for limited expressiveness in their algorithm. Similarly, \cite{jean2018semi} used posterior variance regularization to learn from unlabeled data in situations where labeled data is scarce for improved performance. 

Our work has also strong connections with work on domain adaptation. In particular, estimating ITE requires predictions of outcomes over a different distribution from the observed one. Our ITE error upper bound has similarities with generalization bounds in domain adaptation given by \cite{johansson2016learning,johansson2018learning}. \cite{johansson2018learning} and \cite{zhao2019learning} have similarly argued against enforcing domain-invariance, and related the loss of predictive power of those representations to the loss of information due to the non-invertibility of learned representations. 
\section{ITE Problem formulation}
To estimate what would happen under different interventions, we first frame the causal problem as a statistical one. We build upon the potential outcomes framework \cite{rubin2005causal}, this gives us a formal mathematical representation of the potential outcome using the notation $Y (a)$, which reads as observing $Y$ if we do $a$. 

\textbf{Set up.} Consider a population of individuals with each individual $i$ described by a context $X_i \in \mathcal X$ (typically $\mathbb R^d$). An intervention $T$ is applied to subjects in the population. An important special case is when interventions are binary, $T \in \{0, 1\}$: individual $i$’s response to the intervention is a random variable denoted by $Y_i(1)$, whereas $i$'s natural response when no intervention is applied is denoted by $Y_i(0)$. The two random variables, $Y_i(1), Y_i(0) \in \mathbb R$, are known as the potential outcomes. As discussed, in observational data intervention assignments generally depend on the subjects’ features, i.e. $T_i \notindep X_i$. This dependence is quantified via the conditional distribution $p(T_i | X_i)$, also known as the propensity score of subject $i$.

\textbf{Assumption 1} (Consistency, ignorability and overlap). \textit{For any individual $i$, assigned to intervention $t_i$ , we observe $Y_i = Y(t_i)$. Further, $\{Y (t)\}_{t\in T}$ and the data-generating process $p(X, T, Y )$ satisfy strong ignorability: $Y(0),Y(1) \indep T | X$ and overlap: $\forall x, 0 < p(T | X) < 1$.}

Assumption 1 is a sufficient condition for causal identifiability \cite{rubin2005causal}. Ignorability is also known as the no hidden confounders assumption, indicating that all variables that cause both $T$ and $Y$ are assumed to be measured. Under ignorability therefore, any domain shift in $p(X)$ cannot be due to variables that causally influence $T$ and $Y$, other than through $X$. Under Assumption 1, potential outcomes equal conditional expectations: $\mathbb{E}[Y (t) | X = x] = \mathbb{E}[Y | X = x, T = t]$, and we may predict $Y (t)$ by regression. 

\textbf{Objective.} We attempt to learn predictors $f_t:\mathcal X \rightarrow \mathcal Y$ such that $f_t(x)$ approximates $\mathbb E[Y | X = x, T = t]$. Predictors $f_t = w_t \circ \phi$ are formed by composing a function $w_t:\mathcal Z \rightarrow \mathcal Y$ that operates in a feature space $\mathcal Z$ defined by a representation $\phi:\mathcal X \rightarrow Z$.

The individual effect of an intervention $T = 1$ in context $X$ is measured by the conditional average treatment effect (CATE), $\tau(x) = \mathbb E [Y (1) - Y (0) | X = x]$, and the error in estimation is given by the empirical precision in estimating heterogeneous effects $\epsilon_{\textsc{PEHE}}$, defined as the mean squared error in the estimation of the treatment effect $\tau(x)$,
\begin{align}
\label{pehe}
    \epsilon_{\textsc{PEHE}}= \int \left(\hat \tau(x) - \tau(x)\right)^2 p(x) dx
\end{align}
Predicting $\tau$ for unobserved units involves prediction of both potential outcomes, but since we only observe the "factual" outcome for a specific treatment assignment, and never observe the corresponding "counterfactual" outcome, we never observe any samples of the true treatment effect in an observational dataset. This makes the problem of causal inference fundamentally different from standard supervised learning. 

\textbf{Notation.} We analyse the generalization properties of predictors $f_t$ and underlying representations $\phi$ with respect to $\epsilon_{\textsc{PEHE}}$ in the next section. To do so we will make use of the following notation. Let $\mathcal{P}_{x,t}$ denote the input data distribution $p(x,t)$, $\mathcal{P}_{t}=p(y_t|x)p(x,t)$ the joint factual distribution of $x$ and $y_t$, $\mathcal{P}_{1-t}=p(y_t|x)p(x,1-t)$ the joint counterfactual distribution of $x$ and $y_t$. Each instance in the observed (factual) dataset $\mathcal{D}_{t} = \{(x_{i,t}, y_{i,t})\}_{i=1}^{N_t}$, is assumed to be sampled $i.i.d$ from $\mathcal{P}_t$. $\mathcal{D}_{t-1}$ will be used to denote the unobserved (counterfactual) data set that results from flipping the treatment assignment for each instance. While we assume $\phi$ to be deterministic, in the following section we let $w_t$ be a vector of weights with prior distribution $\pi_t = \mathcal N(\mathbf{0},\lambda_t^{-1}\mathbf{I})$, $\lambda_t > 0$. In this sense, $\pi_t$ defines the hypothesis space $\mathcal F$ of $f_t$ and we write $\hat{\rho}_t$ for the posterior distribution of $f_t$, itself a random variable. We write $\mu(x_{i}|\mathcal{D}_t,\Theta_t)$ and $\sigma^2(x_{i}|\mathcal{D}_t,\Theta_t)$ for its posterior mean and variance given context $x_i$. $\Theta_t$ includes all hyperparameters (both shared parameters (in $\phi$) and specific parameters to each treatment group).

\section{Intuition and theoretical results}
Inherent to the approach of learning representations for counterfactual inference is that the representation must trade-off between containing predictive information about factual outcomes while mitigating the information content that drives the treatment selection policy to ensure good generalization on counterfactuals. In this section, we make several observations about the deficiencies of enforcing domain invariance for this purpose and propose alternatives based on the posterior counterfactual variance. We start with a simple example that illustrates the inadequacy of distributional distances and the benefits of counterfactual variance minimization.

\begin{figure}[t]
\centering
 \includegraphics[width=0.455\textwidth]{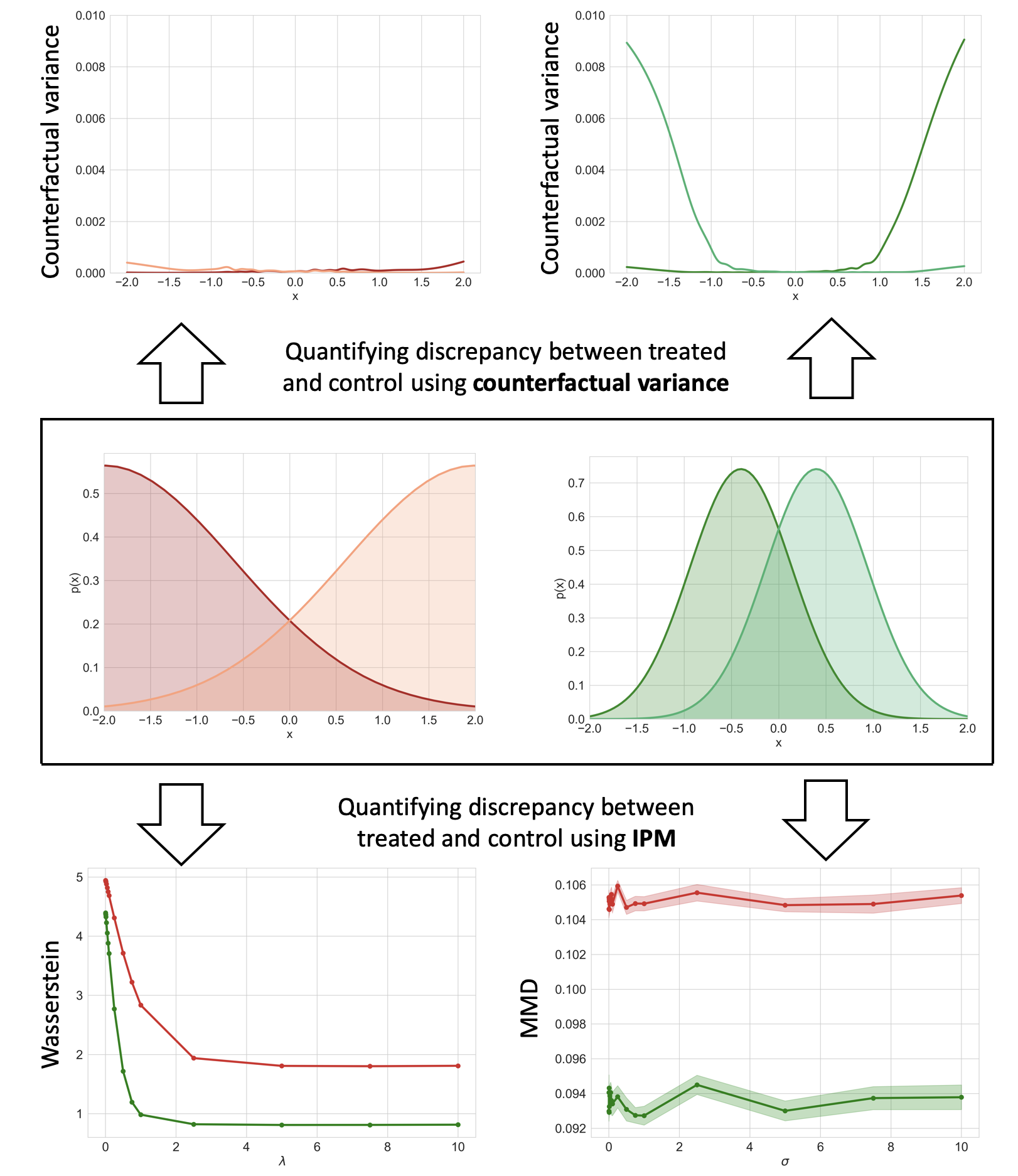}
\caption{Toy example illustrating the shortcomings of distributional distances, like IPMs, for regularizing representations in causal inference. Despite the fact that sufficient support is satisfied in the red population and not in the green population, IPMs (bottom) give the opposite result, with larger discrepancy between groups in the red population than in the green population. In contrast, the counterfactual variance (top) accurately describes the lack of support of the green population.}
\label{fig:toy}
\vspace{-0.5cm}
\end{figure}
\textbf{Toy example} - \textit{Counterfactual variance vs. distributional distances.} In  the middle panels of Figure \ref{fig:toy} we show two simulated datasets.  The left-hand/red dataset arises from truncated normal distributions with large overlap in the tails; the right-hand/green dataset arises from ordinary normal distributions with small overlap in the tails.  In both cases, we show treated and control outcomes in different shades.   In both cases, the outcome is $y=\mathrm{sinc}(4x)$. The red population satisfies sufficient assumptions for identifiability of causal effects; the  green population does not. However, as shown in the bottom panels, both the MMD and Wasserstein distances are {\em smaller} in the green population than in the red population. In contrast, the top panel shows that the predictive variance of the counterfactual outcomes much better describes this lack of overlap. Counterfactual variance is adaptive to the prediction problem of interest (given that it is obtained from a fitted model), providing a data-dependent measure to quantify distances in the underlying function class, perhaps more precise when the underlying function to be estimated are unknown. IPMs are defined as worst-case distances dependent on a function class to be specified a priori. We make the observation also that IPMs need to be approximated in practice which may be inaccurate for high-dimensional representations and small training data samples \cite{boissard2014mean}.

\subsection{Generalization bounds}
In this subsection, we develop a PAC-Bayes generalization bound \cite{mcallester1999some} for the empirical precision in estimating heterogeneous effects $\epsilon_{\textsc{PEHE}}$, that shows specifically why minimizing counterfactual variance can improve generalization performance.

\begin{theorem}\label{thm:thm1} With the assumption that the squared loss function $L:\mathcal{F}\times \mathcal{X}\times\mathcal{Y}\rightarrow\mathbb{R}$ is sub-gaussian under $\pi_t$ and $\mathcal{P}_t$, and using the notation introduced above, the following holds. With probability at least $1-\delta$ and for any posterior distribution $\hat{\rho}_t \ on \ \mathcal{F}$, $\epsilon_{\textsc{PEHE}}$ is upper-bounded by
\begin{align}\label{equ:thm1}
  &\sum_{t=0}^{1}\Bigg(2\tilde{D}_{\infty} L_{\hat{\rho}_t} (\mathbf{X}_t,\mathbf{Y}_t) + (\tilde{D}_{\infty}+1){\Var}_{\hat{\rho}_t}(\mathbf{X}_{t}) \nonumber\\
	  & + {\Var}_{\hat{\rho}_t}(\mathbf{X}_{1-t})+	\frac{C_{t,1}(N_t,N_{1-t})}{N_t N_{1-t} }\KL(\hat{\rho}_t\|\pi_t) \nonumber\\ 
	  &+ \frac{C_{t,2}(N_t,N_{1-t})}{N_t N_{1-t}}\frac{1}{\delta}+C_{t,3} \Bigg)
\end{align}
$C_{t,1}(N_t,N_{1-t})$, $C_{t,2}(N_t,N_{1-t})$ are linear function in their arguments, $C_{t,3}$ is constant, $D_{\infty}(\mathcal{P}_{x,1-t}\|\mathcal{P}_{x,t})={\sup}_{x}\frac{p(x,1-t)}{p(x,t)}$, $L_{\hat{\rho}_t} (\mathbf{X},\mathbf{Y})= \frac{1}{N}\sum_{i=1}^{N} \big(\mu(x_{i}|\mathcal{D}_t,\Theta_t)-y_{i}\big)^2$ is the posterior prediction loss, ${\Var}_{\hat{\rho}_t}(\mathbf{X})$ is the posterior variance on $\mathbf{X}$, ${\Var}_{\hat{\rho}_t}(\mathbf{X})=\frac{1}{N}\sum_{i=1}^{N} \sigma^2(x_{i}|\mathcal{D}_t,\Theta_t)$, $\tilde{D}_{\infty}=D_{\infty}(\mathcal{P}_{x,1-t}\|\mathcal{P}_{x,t})+1$, and finally $\KL(\cdot\|\cdot)$ is the Kullback-Leibler divergence.
\end{theorem}

Observe the dependence on the variance with respect to the distribution of the observed data of the counterfactual outcomes ${\Var}_{\hat{\rho}_t}(\mathbf{X}_{1-t})$.  By minimizing the counterfactual variance we can expect the representations $\phi$ of counterfactual data to encode a relatively smoother prediction function $w \circ \phi (x)$, as can be seen in Figure \ref{fig:embedding}. The estimation of treatment effects is inherently a label-scarce problem as counterfactual data is not observed, representations resulting in smooth prediction curves are especially important to generalize beyond the factual data. This view of the problem of estimating treatment effects emphasizes the need of regularization for good generalization. We note that the sub-gaussian assumption on the loss function may not hold in some cases e.g. large noise. In this case, we may relax this assumption to sub-exponential with all results and proofs otherwise unchanged \cite{germain2016pac}.

% \begin{figure}[H]
% \centering
%  \includegraphics[scale = 0.32]{figure/embedding/tsne_var_control.png}
% \caption{T-SNE visualizations of the learned embeddings for the control potential outcomes $Y(0)$ of the IHDP dataset. A Smooth representation is obtained by optimizing Gaussian likelihood and counterfactual variance jointly. Similar plots optimized for Gaussian likelihood with and without Wasserstein distance can be found in Appendix 4.2.}
% \label{fig:tsne_var_control}
% \end{figure}

% \begin{figure}[H]
% \centering
% 	\includegraphics[scale=0.21]{figure/double_dist_1}
% 	\caption{The mismatch between $\mathcal{P}_{\mathbf{x},1-t}$ and $\mathcal{P}_{\mathbf{x},t}$.}
% \label{fig:double_distribution}
% \end{figure}
% \begin{figure}[H]
% \centering
% 	\includegraphics[scale=0.21]{figure/two_curve2}
% 	\caption{On some input samples from the distribution $\mathcal{P}_{\mathbf{x},t}$ in Figure \ref{fig:double_distribution}, the function $f_{t,A}$ fitting to the observed outcomes in green has larger variance than $f_{t,B}$ fitting to observed outcomes in brown.}
% \label{fig:two_curve}
% \end{figure}

\subsubsection{Why distributional distances may be inadequate?} 
While the toy example clearly illustrates the inability of distributional distances to capture domain overlap, we argue that by enforcing equality in full marginals, optimizing for distributional distances may also overly penalize the model's ability to predict factual observations when sufficient data is available, that is, when domain overlap is satisfied. 

In the following Theorem, we provide a bound on the generalization error on counterfactual data that illustrates the interplay between distribution mismatch and prediction loss on factual data. We show that distribution mismatch becomes decreasingly relevant with increasing data set size. %Note that the term ${\IPM}_\mathcal{H}(\mathcal{P}_{\mathbf{x},1-t},\mathcal{P}_{\mathbf{x},t} )$ in Equation \ref{equ:thm2_1} of Theorem \ref{thm:thm2} is derived using Lemma A4 in \cite{shalit2017estimating} which inspires the recent trend of removing selection bias via IPM minimization.

%  let additionally $\mathcal{H}$ be a family of functions $h:\mathcal{X}\rightarrow \mathbb{R}$, and denote by ${\IPM}_\mathcal{H}(\cdot,\cdot)$ the integral probability metric induced by $\mathcal{H}$. 

\begin{theorem}\label{thm:thm2}
Assume the notation introduced above, for any posterior distribution $\hat{\rho}_t \ on \ \mathcal{F}$, the expected counterfactual Gibbs risk\footnote{The expected counterfactual Gibbs risk is an upper bound of the expected Bayesian counterfactual risk, a detailed discussion of Gibbs and Bayesian risk is given in Appendix 1.2.} $R_{\mathcal{P}_{1-t}}(G_{\hat{\rho}_t})$ is bounded above by,	
\begin{equation}\label{equ:thm2}
		\frac{1}{2}D_{\mathcal{P}_{x,1-t}}(G_{\hat{\rho}_t}) + D_{\infty}(\mathcal{P}_{x,1-t}\|\mathcal{P}_{x,t})L_{\mathcal{P}_{t}}(G_{\hat{\rho}_t}).
%		+\eta_{\mathcal{P}_{1-t}/\mathcal{P}_{t}}
	\end{equation}
%Furthermore, assuming there exists $\gamma>0$ such that the expectation $h_{f,f',t}(x)=\frac{1}{\gamma}\mathbb{E}_{(f,f')\sim \hat{\rho}_t^2}\mathbb{E}_{y_t\sim p(y_t|x)}\quooo{\tilde{l}_{f,f'}(x,y_t)}\in \mathcal{H}$, then $R_{\mathcal{P}_{1-t}}(G_{\hat{\rho}_t})$ is bounded above by,
%\begin{equation}\label{equ:thm2_1}
% \frac{1}{2}D_{\mathcal{P}_{x,1-t}}(G_{\hat{\rho}_t}) + L_{\mathcal{P}_{t}}(G_{\hat{\rho}_t})+ \gamma {\IPM}_\mathcal{H}(\mathcal{P}_{x,1-t},\mathcal{P}_{x,t} )
% \end{equation}
where $D_{\mathcal{P}_{x,1-t}}(\hat{\rho}_t)= \mathbb{E}_{x\sim \mathcal{P}_{x,1-t}} \quo{\sigma^2(x|\mathcal{D}_t,\Theta_t)}$ is the expected counterfactual variance, $L_{\mathcal{P}_t}(\hat{\rho}_t)=  \mathbb{E}_{(x,y)\sim \mathcal{P}_t} \quooo{\big(\mu(x|\mathcal{D}_t,\Theta_t)-y\big)^2 +\frac{1}{2}\sigma^2(x|\mathcal{D}_t,\Theta_t)}$ is the expected factual loss, and finally $D_{\infty}(\mathcal{P}_{x,1-t}\|\mathcal{P}_{x,t})=\sup_{x}\frac{p(x,1-t)}{p(x,t)}$.
\end{theorem}

The bound in Equation (\ref{equ:thm2}) describes the interaction between the distribution mismatch and the prediction error on factual data $L_{\mathcal{P}_{t}}(G_{\hat{\rho}_t})$. $D_{\infty}(\mathcal{P}_{x,1-t}\|\mathcal{P}_{x,t})$ is large if, for some $x$, $\mathcal{P}_{x,t}$ has small density while $\mathcal{P}_{x,1-t}$ has large density (that is when there is poor overlap between them), which understandably, makes minimizing the counterfactual risk $R_{\mathcal{P}_{1-t}}(G_{\hat{\rho}_t})$ harder because few examples from the other population are observed for a given context. However, note that $D_{\infty}(\mathcal{P}_{x,1-t}\|\mathcal{P}_{x,t})$ is multiplied by the expected factual loss $L_{\mathcal{P}_{t}}(G_{\hat{\rho}_t})$ (which decreases as the number of factual samples increases, see Equation (10) in Theorem 7 of Appendix 1.3). The distribution mismatch $D_{\infty}(\mathcal{P}_{x,1-t}\|\mathcal{P}_{x,t})$ thus becomes less important for generalization, if we can minimize the expected factual loss $L_{\mathcal{P}_{t}}(G_{\hat{\rho}_t})$ arbitrarily well. This suggests that optimizing for distributional distances between treated and control data \textit{at the expense} of prediction error on the factual data may be counterproductive. Representations regularized with distributional distances may thus shrink the hypothesis space and converge on solutions that, although balanced across treatment groups, loose their predictive power.

Here again, we note the presence of the counterfactual variance $L_{\mathcal{P}_t}(\hat{\rho}_t)$ in equation (\ref{equ:thm2}) which is not considered by methods optimizing for the distributional difference only.

\subsection{Why encourage preserving information content?}
Assumption 1 gives sufficient conditions for unbiased estimation of the treatment effect using observational data, but identifiability need not hold with respect to the feature representation $Z=\phi(X)$, even if it does with respect to $X$. For instance consider $\phi^{-1}(z):=\{x: \phi(x) = z\}$, then,
\begin{align}
    p(Y|\phi(x),t) &= \frac{\int_{x \in \phi^{-1}(z)}p(Y|x,t)p(x|t)}{\int_{x \in \phi^{-1}(z)}p(x|t)} \nonumber\\
    &=\frac{\int_{x \in \phi^{-1}(z)}p(Y|x)p(x|t)}{\int_{x \in \phi^{-1}(z)}p(x|t)} \nonumber\\
    &\neq p(Y|\phi(x))
\end{align}
in general, with equality only if $\phi$ is invertible. The conditional independence in the strong ignorability assumption, $Y(0),Y(1)\indep T | \phi(X)$, required for estimating the treatment effect need not hold for non-invertible transformations. In this sense, we may be introducing unobserved confounders in representation space we hypothesize by the information lost in the map $\phi$. Observe also that our objective, the conditional average treatment effect $\tau(x) = \mathbb{E}[Y (1) - Y (0) | X = x]$, is expressed in terms of expectations. Similarly, it holds that in feature space, $\mathbb{E}[Y (1) - Y (0)|\phi(x)] = \int_{x \in \phi^{-1}(z)} \mathbb{E}[Y (1) - Y (0)|x] dx$ will in general \textit{not} be equal to our quantity of interest $\tau(x) = \mathbb{E}[Y (1) - Y (0)|x]$ unless $\phi$ is invertible. 

\section{Algorithms}
In this section, we describe a model for counterfactual estimation, called DKLITE\footnote{An implementation of DKLITE is available at https://bitbucket.org/mvdschaar/mlforhealthlabpub.} (Deep Kernel Learning for Individualized Treatment Effects), motivated by our analysis. Our proposed prediction algorithm works in a feature space induced by $\phi$ indirectly via a kernel function $K:\mathcal X \times \mathcal X \rightarrow \mathbb R$, that intuitively models the correlation between inputs $x, x'\in\mathcal X$ in a possibly high-dimensional feature space. We extend the expressiveness of $K$ by transforming the inputs through a non-linear mapping $\phi : \mathcal X \rightarrow \mathbb{R}^{d_{\phi}}$ to form $k(\phi(x), \phi(x'))$, a \textit{deep kernel} \cite{wilson2016deep}. The dimension $d_{\phi}$ of $\phi(x)$ can be chosen arbitrarily, and may therefore differ from the dimension of $x$. We let $\phi_{\theta}$ be parameterized by a neural network with parameters $\theta$ to encode a the information content of input variables. 

\textbf{Prediction model.} The potential outcomes given individuals covariates are assumed to take the form,
\begin{equation}\label{equ:white}
 	y_{i,t}=f_t(x_i)+\epsilon_{i,t}, \qquad t\in\{0,1\}
 \end{equation}
where $f_t(x_i) = w_t^{\top}\mathbf{\phi}(x_i)$ is given as a linear combination of feature representations $\mathbf{\phi}(\cdot)$. $w_t \in \mathbb R^d$ is a weight vector, $\mathbf{\phi}(\cdot)$ the representation layer  and $\epsilon_{i,t}\sim\mathcal N(0,\beta_t^{-1})$ is a noise variable, $t=0,1$. We specify our uncertainty in parameter values through the specification of prior distributions. Combined with the likelihood of observed data, these determine the posterior distribution of parameters and predictive distributions of treatment effects. For each component of the weight vector, independently, we assume $w_t\sim\mathcal N(0,\lambda_t^{-1})$. With knowledge of $\phi$, the posterior of $w_t$ is given by,

\begin{equation}\label{equ:mean_var}
	p(w_t|\mathcal{D}_t,\Theta_t)=\mathcal{N}(\mathbf{m}_{w_t},\mathbf{K}_{w_t}^{-1})
\end{equation}
where $\Theta_t = \{\phi,\beta_t,\lambda_t \}$, the posterior mean $\mathbf{m}_{w_t} = \beta_t\mathbf{K}_{w_t}^{-1}\mathbf{\Phi}_t^{\top}\mathbf{Y}_t$, the posterior covariance $\mathbf{K}_{w_t}^{-1}= \beta_t\mathbf{\Phi}_t^{\top}\mathbf{\Phi}_t+\lambda_t\mathbf{I}_{d_{\phi}\times d_{\phi}}
$ and $\mathbf{\Phi}_t$ the representation of $\mathbf{X}_t$. However, it can be difficult to encode our knowledge about good representation spaces for counterfactual estimation in a Bayesian prior. Posterior regularization offers a more direct and flexible mechanism for guiding and controlling the posterior distribution.

\textbf{Regularized Bayes.} The regularized posterior is the solution of the following optimization problem \cite{zhu2014bayesian,jean2018semi}:
\begin{align}\label{equ:regbayes}
\inf_{q(f|\mathcal{D})}&\mathcal{L}(q(f|\mathcal{D}))+\Omega(q(f|\mathcal{D}))
\end{align}
where $\mathcal{L}(q(f|\mathcal{D}))$ is the (KL-divergence between the approximate posterior $q(f|\mathcal{D})$ and the true posterior $p(f|\mathcal{D})$, and $\Omega(q(f|\mathcal{D}))$ is a regularizer of the approximate posterior $q(f|\mathcal{D})$. We attempt therefore to learn a posterior distribution as close as possible to the true posterior while also fulfilling the requirements imposed by the regularization term.

\textbf{Inference.} The predictive distribution for a testing point $x$ is obtained by marginalising over the posterior $w_t$. It is given by
\begin{equation}\label{equ:pred_p1}
f_t(x) \sim \mathcal{N}\big(\mu(x|\mathcal{D}_t,\Theta_t),\sigma^2(x|\mathcal{D}_t,\Theta_t)\big)	
\end{equation}
where $\mu(x|\mathcal{D}_t,\Theta_t)=\mathbf{m}_{w_t}^{\top}\mathbf{\phi}(x)$ and $\sigma^2(x|\mathcal{D}_t,\Theta_t) = \mathbf{\phi}(x)^{\top} \mathbf{K}_{w_t}^{-1}\mathbf{\phi}(x)$. Point estimates are given for example by the posterior mean or median, optimal for minimizing squared loss or absolute loss, respectively. Note also that knowledge of the full posterior allows us to quantify our uncertainty around point estimates through credible intervals, especially useful in medicine and public policy, for example.
\subsection{Learning $\phi$}
The framework of regularized Bayes described in (\ref{equ:regbayes}) provides us with a flexible optimization objective, many previous algorithms for treatment effects can be reduced to specific instantiations of this problem. In this section we describe the likelihood and regularization terms that we use to encode the intuition and insights derived from section 3. In Appendix 2, we show that the final loss function we derive can be reformulated a specific instance of problem (\ref{equ:regbayes}).

\textbf{Likelihood of the factual data.} Assuming the discriminative model introduced in \ref{equ:white}, we encourage good prediction of the factual data by optimizing for the negative log-likelihood of the factual data,
\begin{align}
\mathcal L_{lik} = -\sum_{t=0}^1 \log(p(\mathbf{Y}_t|\mathbf{X}_t,\Theta_t))
\end{align}
Observe that we may re-write the negative log-likelihood, $\mathcal L_{lik}$, as
\begin{align}\label{equ:mle}
  \mathcal{L}_{lik} = & \sum_{t=0}^1 \Big( \frac{N_t}{2}\ln(2\pi\beta_t^{-1})+
  \KL(\hat{\rho}_t\|\pi_t) \nonumber \\ 
  & +\frac{N_t\beta_t}{2}\big[{\Var}_{\hat{\rho}_t}(\mathbf{X}_{t})+L_{\hat{\rho}_t} (\mathbf{X}_t,\mathbf{Y}_t)\big] \Big) 
\end{align}
From Theorem \ref{thm:thm1}, we see that the empirically estimable quantities to optimize for in the upperbound (\ref{equ:thm1}), are ${\Var}_{\hat{\rho}_t}(\mathbf{X}_{1-t})$, ${\Var}_{\hat{\rho}_t}(\mathbf{X}_{t})$, $L_{\hat{\rho}_t} (\mathbf{X}_t,\mathbf{Y}_t)$ and $\KL(\hat{\rho}_t\|\pi_t)$. The latter three already exist in our objective $\mathcal{L}_{lik}$, as shown in Equation (\ref{equ:mle}). 

{
	\renewcommand{\arraystretch}{1.5}
	\begin{table*}[t]
	\fontsize{7.5}{7.5}\selectfont
		\centering
		\resizebox{\linewidth}{!}{
			\begin{tabular}{|l|c|c|c|c|c|c|}
			
				\hline &
				\multicolumn{2}{c}{IHDP \ ($\sqrt{\hat{\epsilon}_{\textsc{PEHE}}}$)} & \multicolumn{2}{|c|}{Twins \ ($\sqrt{\tilde{\epsilon}_{\textsc{PEHE}}})$} & \multicolumn{2}{c|}{Jobs \  ($\hat{\mathcal{R}}_{pol}(\pi_f)$)}\\
				\hline
				  & In-sample & Out-sample & In-sample & Out-sample & In-sample & Out-sample  \\
				\hline
				\textsc{$\textsc{OLS/LR}_1$} & 5.8 $\pm$ .3 &5.8 $\pm$ .3 &.319 $\pm$ .001 &.318 $\pm$ .007 & .22 $\pm$ .00 & .23 $\pm$ .02    \\
				\hline
				\textsc{$\textsc{OLS/LR}_2$} & 2.4 $\pm$ .1& 2.5 $\pm$ .1 & .320 $\pm$ .002 & .320 $\pm$ .003 & .21 $\pm$ .00 & .24 $\pm$ .01    \\
				\hline
				\textsc{BLR}  & 5.8 $\pm$ .3 & 5.8 $\pm$ .3 & .312 $\pm$ .003 & .323 $\pm$ .018 & .22 $\pm$ .01 & .26 $\pm$ .02    \\
				\hline
				\textsc{k-NN}  & 2.1 $\pm$ .1 & 4.1 $\pm$.2 & .333 $\pm$ .001 & .345 $\pm$ .007 & .22 $\pm$ .00 & .26 $\pm$ .02    \\
				\hline
				\textsc{BART}  & 2.1 $\pm$ .1 & 2.3 $\pm$ .1 & .347 $\pm$ .009 & .338 $\pm$ .016 & .23 $\pm$ .00 & .25 $\pm$ .02    \\
				\hline
				\textsc{R-Forest}  & 4.2 $\pm$ .2& 6.6 $\pm$ .3 & .366 $\pm$ .002 & .321 $\pm$ .005 & .23 $\pm$ .01 & .28 $\pm$ .02    \\
				\hline
				\textsc{C-Forest}  & 3.8 $\pm$ .2 & 3.8 $\pm$.2  & .366 $\pm$ .003 & .316 $\pm$ .011 &  .19 $\pm$ .00 & .20 $\pm$ .02    \\				
				\hline
				\textsc{BNN}  & 2.2 $\pm$ .1 & 2.1 $\pm$ .1 & .325 $\pm$ .003 & .321 $\pm$ .018 &  .20 $\pm$ .01 & .24 $\pm$ .02    \\
				\hline
				\textsc{TARNET}  & .88 $\pm$.02 & .95 $\pm$ .02  & .317 $\pm$ .002 & .315 $\pm$ .003 &  .17 $\pm$ .01 & .21 $\pm$ .01    \\
				\hline
				\textsc{$\textsc{CAR}_{\textsc{WASS}}$}  & .72 $\pm$ .02 & .76 $\pm$ .02 & .315 $\pm$ .007  & .313 $\pm$ .008  &  .17 $\pm$ .01  & .21 $\pm$ .01    \\
				\hline
				\textsc{CMGP}  & .63 $\pm$ .08 & .74 $\pm$ .11  & .320 $\pm$ .002 & .319 $\pm$ .008  & .22 $\pm$ .03 & .24 $\pm$ .05    \\
				\hline
				\textbf{\textsc{DKLITE}}  & \textbf{.52 $\pm$ .02} & \textbf{ .65 $\pm$ .03 } & \textbf{ .288 $\pm$ .001} & \textbf{ .293 $\pm$ .003} & \textbf{ .13 $\pm$ .01} & \textbf{ .14 $\pm$ .01 }  \\
				\hline
			\end{tabular}}
			\caption{Mean performance (lower better) of individualized treatment effect estimation and standard deviation.}
			\vspace{-0.44cm}
			\label{tab:realdata1}
	\end{table*}
}
\textbf{Posterior variance as regularization.}
Therefore, by including an empirical estimate of the counterfactual variance ${\Var}_{\hat{\rho}_t}(\mathbf{X}_{1-t})$ as a regularizer in our objective we are effectively optimizing for all terms given by the PAC-Bayes upperbound in equation (\ref{equ:thm1}). We write this regularization term as,
\begin{align}
    \mathcal L_{var} = \sum_{t=0}^1 \frac{1}{N_{1-t}} \sum_{i=1}^{N_{1-t}}\sigma^2(x_{i,1-t}|\mathcal{D}_t,\Theta_t)
\end{align}
Since the deep kernel parameters are jointly learned, the neural net $\phi$ is encouraged to learn a feature representation in which the counterfactual examples are close to the factual examples, thereby reducing the variance on our predictions. The implication is that we are optimizing for representations where counterfactual data tend to cluster around the representations of factual data. This is a way to see how intuitively that we are encouraging overlap in support in representation space \textit{without} enforcing equality in densities (i.e. the size of the factual and counterfactual "clusters" need not coincide). 

\textbf{Invertibility as regularization.}
While the loss due to non-invertible representations is not directly observable, we may associate it with the information content of $x$ lost in $\phi(x)$ and we found it to be an important source of gain in performance, empirically. We can encourage information content preservation with an additional decoder $\psi:\mathbb R^{d_{\phi}} \rightarrow \mathcal X$ - a neural network with the reversed structure of network $\phi$ (the encoder in this sense) - trained to reconstruct the input $x$ from $\phi(x)$. The reconstruction loss is given as,
\begin{align}
    \mathcal L_{rec} = \sum_{t=0}^1 \frac{1}{N_t}\sum_{i=1}^{N_t}\|x_{i,t}-\mathbf{\psi}(\mathbf{\phi}(x_{i,t}))\|_{2}^{2}
\end{align}
We remark that some recent density estimation approaches, e.g. normalization flows \cite{rezende2015variational} and non-volume preserving transformation \cite{dinh2016density}, can be used as an alternative method to achieve invertibility. 

\textbf{Final loss function.}  Based on the objectives described above, our final loss trades-off between maximizing the likelihood of the observed (factual) data under our model, minimizing the predictive variance of the counterfactual outcomes and minimizing the reconstruction loss of the representations. The loss is given as follows, 
\begin{align}
    \mathcal L_{fin} = \mathcal L_{lik} + \alpha_1 \mathcal L_{var} + \alpha_2 \mathcal L_{rec}
\end{align}
$\alpha_1>0$ and $\alpha_2>0$ are hyperparameter. Standard methods for hyperparameter selection are not generally applicable for choosing hyperparameters because counterfactuals are never observed. As an approximation scheme, we replace the missing counterfactuals with their nearest factual neighbour to compute the treatment effects in cross-validation \cite{shalit2017estimating}. 
\section{Experiments}
The primary focus of our experiments will be on the comparison of prediction performance on benchmark tasks, the use of the posterior variance for decision-making and, a deeper analysis of learned representations and source of performance gain. In the following, we start by introducing competing prediction algorithms before giving a brief description of the data and appropriate performance metrics.

\textbf{Competing algorithms.} We compare DKLITE with a total of 11 algorithms. First we evaluate least squares regression using treatment as an additional input feature $\textsc{OLS/LR}_1$), we consider separate least squares regressions for each treatment ($\textsc{OLS/LR}_2$), we evaluate balancing linear regression (BLR) \cite{johansson2016learning}, k-nearest neighbor (k-NN) \cite{crump2008nonparametric}, Bayesian additive regression trees (BART) \cite{chipman2010bart}, random forests (R-Forest) \cite{breiman2001random}, causal forests (C-Forest) \cite{wager2018estimation}, balancing neural networks (BNN) \cite{johansson2016learning}, treatment-agnostic representation network (TARNET), counterfactual regression with Wasserstein distance $(\textsc{CAR}_{\textsc{WASS}})$ \cite{shalit2017estimating}, and multi-task gaussian process (CMGP) \cite{alaa2017bayesian}.

\textbf{Data.} Causal inference models are often impossible to reliably validate using real-world data due to the absence of counterfactual outcomes. Various established approaches for evaluating causal models have been proposed, which we use for our analysis. We describe these briefly below and refer the reader to the accompanying references and section 5 of Appendix for further details. Additional results (e.g. average treatment effect estimation) are included in the same Appendix section. We consider \textbf{IHDP} \textit{( 747 instances described by 25 covariates)} \cite{hill2011bayesian,alaa2017bayesian,shalit2017estimating, yao2018representation} in which counterfactual outcomes are randomly generated via a predefined probabilistic model; \textbf{Twins} \textit{(11300 instances described by 30 covariates)}, in which outcomes are observed but the treatment assignment is simulated; and finally, \textbf{Jobs} \textit{(3212 instances described by 7 covariates)} \cite{shalit2017estimating,lalonde1986evaluating,dehejia2002propensity} which is constructed from a mixture of experimental and observational data with the treatment. 

\textbf{Metrics.} The metrics used to evaluate each data set differ slightly depending on the available outcome (real or simulated) and the available treatment assignment mechanisms (known or unknown). For the \textbf{IHDP}, we use the empirical precision in estimating treatment effects $\hat{\epsilon}_{\textsc{PEHE}}=\frac{1}{N}\sum_{i=1}^{N}(\tau(x_i)-\hat{\tau}(x_i))^2$. For \textbf{Twins}, we use the observed precision in estimating heterogeneous effects, $\tilde{\epsilon}_{\textsc{PEHE}}=\frac{1}{N}\sum_{i=1}^{N}(y_{i,1}-y_{i,0}-\hat{\tau}(x_i))^2$. For Jobs, we use the policy risk that measures the expected loss if the treatment is taken according to the ITE policy prescribed by the algorithm, $\mathcal R_{pol} = 1-\mathbb E[Y(1)|\pi(x) = 1]P(\pi(x) = 1)+\mathbb E[Y(0)|\pi(x) = 0]P(\pi(x) = 0)$ where $\pi(x) = 1$ if $\hat y(1)-\hat y(0) > 0$ and $\pi(x) = 0$, otherwise.

\subsection{Prediction performance results}
We report in-sample and out-of-sample performance in Table \ref{tab:realdata1}. DKLITE targets aspects of the treatment effect estimation problem that have not been considered before. Learning with such an objective significantly outperforms all competing algorithms and does so on all benchmark data sets. The most relevant comparison is perhaps with $\textsc{BNN}$ \cite{johansson2016learning} and $\textsc{CAR}_{\textsc{WASS}}$ \cite{shalit2017estimating}, neural network models that enforces domain invariance through distributional distances. The performance gain highlights the predictive power of our representations.

\subsection{Source of gain}

\newcolumntype{G}{>{\centering\arraybackslash}m{1.6cm}}
\newcolumntype{H}{>{\centering\arraybackslash}m{1.6cm}}
\newcolumntype{I}{>{\centering\arraybackslash}m{1.6cm}}
\newcolumntype{J}{>{\centering\arraybackslash}m{1.6cm}}

\begin{table}[H]
\fontsize{7.5}{8.5}\selectfont
\centering
\begin{tabular}{|G|HIJ|}
 \hline
    & $\mathcal L_{lik}$ & $\mathcal L_{lik} + \mathcal L_{var}$ & $\mathcal L_{lik} +\mathcal L_{rec}$ \\
 \hline
 \textbf{IHDP}   &    & & \\
 In-sample   &  1.46 $\pm$ .01   & .98 $\pm$ .04 & .96 $\pm$ .04\\
 Out-sample &  1.95 $\pm$ .14   & 1.24 $\pm$ .09  & 1.28 $\pm$ .13\\
 \textbf{Twins}   &    &  &\\
 In-sample   &  .308 $\pm$ .004   & .292 $\pm$ .002 &.291 $\pm$ .001\\
 Out-sample &  .323 $\pm$ .008   & .294 $\pm$ .007  &.293 $\pm$ .003\\
 \textbf{Jobs}   &    &  &\\
 In-sample   &  .18 $\pm$ .01   & .14 $\pm$ .01 & .15 $\pm$ .01\\
 Out-sample &  .28 $\pm$ .01   & .23 $\pm$ .01  & .24 $\pm$ .01\\
 \hline
\end{tabular}
\caption{Source of performance gain. }
\label{gain}
\vspace{-0.4cm}
\end{table}

In this section, we analyze more deeply the contribution to the performance gain of each component of our loss. We evaluate DKLITE optimized for different components of $\mathcal L_{fin} = \mathcal L_{lik} + \alpha_1 \mathcal L_{var} + \alpha_2 \mathcal L_{rec}$. As can be seen in Table \ref{gain}, including regularization based on the counterfactual variance ($\mathcal L_{lik} + \alpha_1 \mathcal L_{var}$) and reconstruction loss ($\mathcal L_{lik} + \alpha_2 \mathcal L_{rec}$), each evaluated separately, already provides a significant gain in performance with respect to optimization on the factual data only ($\mathcal L_{lik}$). Importantly though, combining them ($\mathcal L_{fin}$) improves performance further by an order of magnitude (see DKLITE in Table \ref{tab:realdata1}), which suggests that $\mathcal L_{var}$ and $\mathcal L_{rec}$ capture to some extent "orthogonal" sources of gain. The gain is especially important on relatively smaller data sets, such as IHDP with 747 individuals, and to a lesser extent on bigger data sets. These results illustrate the behaviour suggested by Equation \ref{equ:thm2} in Theorem \ref{thm:thm2}: distribution mismatch between groups becomes decreasingly relevant with increasing data set size. In this setting, the error on the factual data ($\mathcal L_{lik}$) drives generalization performance.

\subsection{Leveraging the predicted uncertainty }
Data-driven solutions for decision support have most often been proposed without methods to quantify and control their uncertainty in a decision. In contrast, in medicine for example, a physician knows whether she is uncertain about a case and will consult more experienced colleagues if needed. We use this idea to show that uncertainty informed treatment effect estimation can improve performance. An instantiation of this approach, termed DKLITE-U, is given by referring the 10 $\%$ most uncertain predictions for further scrutiny. Performance in comparison to DKLITE is given in Table \ref{DKLITE-U}. Note that especially on small data sets, such as IHDP, predictions can be significantly improved.

\newcolumntype{K}{>{\centering\arraybackslash}m{1.8cm}}
\newcolumntype{L}{>{\centering\arraybackslash}m{1.6cm}}
\newcolumntype{M}{>{\centering\arraybackslash}m{1.6cm}}

\begin{table}[H]
\fontsize{7.5}{8.5}\selectfont
\centering
\begin{tabular}{ |K|L|M|  }
 \hline
    & \textbf{DKLITE} & \textbf{DKLITE-U} \\
 \hline
 \textbf{IHDP}   &    &  \\
 In-sample   &  .52 $\pm$ .02   & .46 $\pm$ .02\\
 Out-sample &  .60 $\pm$ .03   & .53 $\pm$ .02  \\
 \textbf{Twins}   &    &  \\
 In-sample   &  .288 $\pm$ .001   & .287 $\pm$ .001\\
 Out-sample &  .293 $\pm$ .003   & .292 $\pm$ .002  \\
 \textbf{Jobs}   &    &  \\
 In-sample   &  .13 $\pm$ .01   & .13 $\pm$ .01\\
 Out-sample &  .14 $\pm$ .01   & .14 $\pm$ .01  \\
 \hline
\end{tabular}
\caption{Performance of DKLITE-U on all data sets. }
\label{DKLITE-U}
\vspace{-0.4cm}
\end{table}

\section{Discussion}
In many domains understanding the effect of interventions at an individual level is crucial, but prediction of those potential outcomes is challenging. Despite their empirical success, we find that algorithms enforcing representations to satisfy domain invariance is often too strong a requirement for causal predictions. This stems from the fact that overlapping support is sufficient for identifiability of the causal effect and equality in densities is not necessary. 

We have proposed a bound on generalization performance that shows the dependence on domain overlap through the counterfactual variance which we interpret as a proxy for domain overlap, and highlighted the need for invertible latent maps. These results motivated novel regularization criteria that we incorporated in a deep kernel learning framework through posterior regularization. We hypothesize that many existing models, both frequentist (through representer theorems if optimized with empirical risk minimization) and Bayesian, can be reduced to specific instances of our framework with different regularization terms. We leave a formal derivation of this unifying theory for future work.

%%%%%%%%%%%%%%%%%%%%%%%%%%%%%%%%%%%%%%%%%%%%%%%%%%%%%%%%%%%%%%%%%%%%%%%%%%%%%%%%%%%%%%%%%%%%%%%%%%
\section*{Acknowledgements}
%%%%%%%%%%%%%%%%%%%%%%%%%%%%%%%%%%%%%%%%%%%%%%%%%%%%%%%%%%%%%%%%%%%%%%%%%%%%%%%%%%%%%%%%%%%%%%%%%%
We thank the anonymous reviewers for valuable feedback. This work was supported by GlaxoSmithKline (GSK), the Alan Turing Institute under the EPSRC grant EP/N510129/1, the US Office of Naval Research (ONR), and the National Science Foundation (NSF): grant numbers ECCS1462245, ECCS1533983, and ECCS1407712.
\balance
%\clearpage
%\newpage
\bibliography{sample_paper}
\bibliographystyle{IEEE}

\end{document}